\documentclass[runningheads,a4paper]{llncs}

\usepackage{amssymb}
\usepackage{graphicx,subfigure,amsmath}
\usepackage{url}
\urldef{\mailsa}\path|{alfred.hofmann, ursula.barth, ingrid.haas, frank.holzwarth,|
\urldef{\mailsb}\path|anna.kramer, leonie.kunz, christine.reiss, nicole.sator,|
\urldef{\mailsc}\path|erika.siebert-cole, peter.strasser, lncs}@springer.com|
\newcommand{\keywords}[1]{\par\addvspace\baselineskip
\noindent\keywordname\enspace\ignorespaces#1}

\begin{document}

\mainmatter  
\newcommand{\var}[1]{\mathbf{#1}}
\newcommand{\set}[1]{\mathcal{#1}}
\title{Saliency Supervision: An intuitive and effective approach for pain intensity regression}

\titlerunning{Saliency Supervision for Pain Intensity Regression}

%
%
\author{Conghui Li$^1$, Zhaocheng Zhu$^2$, Yuming Zhao$^{1}$\thanks{Corresponding author (e-mail: arola\_zym@sjtu.edu.cn)}}
\authorrunning{Conghui Li, Zhaocheng Zhu,Yuming Zhao}

\institute{
School of Electronic Information and Electrical Engineering\\
Shanghai Jiao Tong University$^1$\\
School of Electronics Engineering and Computer Science\\
Peking University$^2$\\
}

%
%

\toctitle{Lecture Notes in Computer Science}
\tocauthor{Authors' Instructions}
\maketitle

\begin{abstract}
Getting pain intensity from face images is an important problem in autonomous nursing systems. However, due to the limitation in data sources and the subjectiveness in pain intensity values, it is hard to adopt modern deep neural networks for this problem without domain-specific auxiliary design. Inspired by human vision priori, we propose a novel approach called saliency supervision, where we directly regularize deep networks to focus on facial area that is discriminative for pain regression. Through alternative training between saliency supervision and global loss, our method can learn sparse and robust features, which is proved helpful for pain intensity regression. We verified saliency supervision with face-verification network backbone \cite{Wen2016A} on the widely-used \textit{UNBC-McMaster Shoulder-Pain} \cite{5771462} dataset, and achieved state-of-art performance without bells and whistles. Our saliency supervision is intuitive in spirit, yet effective in performance. We believe such saliency supervision is essential in dealing with ill-posed datasets, and has potential in a wide range of vision tasks.
\end{abstract}

\section{Introduction}

Excessive usage of anesthetic will cause bad effects on patinets because the pain intensity of the patient is not well measured by tradition methods like skin conductance algesimeter or heart rate variability, which calls for a reliable approach that reports pain intensity in time. From the perspective of computer vision, the pain intensity task can be viewed as a semantic problem from single images or videos. Our job is aimed at pushing the benchmark of pain intensity to state-of-art level with a new design.

There are a few datasets that have pain intensity labels. Particularly, \textit{UNBC-McMaster Shoulder-Pain} \cite{5771462}  dataset is the only dataset available for per-frame visual analysis. It contains only 200 videos of 25 patients who suffer from shoulder pain and repeatedly raise their arms and then put them down (onset-apex-offset). While all frames are labeled with pain intensities, the labels are reported by the patient, which is very subjective. Moreover, in most frames patients got zero pain intensity, making it hard to observe the pattern behind frames. Many methods \cite{Wang2017Regularizing} \cite{Chang2017FATAUVA} have exploited deep neural networks trained in a data-extensive domain to alleviate the limited training data problem. For example, they finetune a well-trained face verification network with a regularized regression loss for pain intensity regression or valence-arousal estimation. In this case, their initial value is expected to be closer to some optimum for pain intensity regression, due to the similarity between domains. However, we doubt the finetuning procedure in those methods, as pain intensity values is not always a good supervision signal, especially for small dataset like \textit{UNBC-McMaster Shoulder-Pain}.

 Many people would give their reasons based on face attributes like eyes or lips when asked to determine individual's pain intensity from his face. Inspired by this fact, we exploit Action Units (AUs) \cite{Ekman1997What}, a semantic representation of face attributes, to regularize the training of pain intensity regression. AUs have been utilized for facial expression recognition. For example, FATAUVA-Net \cite{Chang2017FATAUVA} train a mid-level network for AU detection using the AU label, and then finetune mid-level network for valence-arousal estimation and facial expression recognition. They use the labeled AU values as additional information and get better result. Pain is also a kind of expression, and here we explicitly supervise the network to focuse on the areas of AUs so that the network will pay attention to the areas related to pain which will give a closer initial value to the global optimum than the previous work \cite{Wang2017Regularizing} when finetuning the network for pain intensity regression.

Since deep neural networks work like a black box, we cannot exactly know where the network pay attention. Zeiler proposed a kind of visualizing network method \cite{Matthew2013Visualizing} using deconvolution \cite{Zeiler2010Deconvolutional} to get the contribution of input to the network's output. As we want to directly supervise the saliency map, which means the network for getting saliency should be able to merge with the origin network and support forward and backward calculation. Deconvolution is a good option for generating saliency map for our task. We add a deconvolution group after the bottleneck layer, making the network an encoder-decoder, note that in this encoder-decoder framework, the framework is different from the task of image generation \cite{Radford2015Unsupervised} or image completion \cite{DBLP:journals/corr/YehCLHD16}, the deconvolution group shares the architecture and weights with convolution group, and we hope the decoded saliency map is similar with attention map which means the saliency map observed from the bottleneck layer should follow certain distribution where some areas such as mouth, eyes should be more activated than others.

According to Facial Action Coding System, six common emotions: happiness, sadness, surprise, fear, anger, disgust, contempt are coded by the AU. Since not all AUs are related to a specific expression, we can not directly use the AU value as supervised signal.  To avoid the intra-variance AUs that may case wrong supervision,  we organize the inputs in the form of triplets \cite{DBLP:journals/corr/SchroffKP15} to use the relative value of pain intensity. In a triplet, the anchor and positive have the same pain intensity, and the negative has different pain intensity. If an AU is almost the same between anchor and positive and different between anchor and negative, we think such AU is related to pain, otherwise it belongs to irrelevant AUs. We select possible related AUs by comparing the pain intensity and AU labels in every triplet and then build the attention map for the triplet. At the same time, using triplets as input for saliency supervision, we alleviate the imbalance problem of training data through utilizing most image instances whose pain intensity is zero and avoiding overfitting.

As for the encoder-decoder framework, the deconvolution group is used for saliency map generation while not for feature learning, so we freeze the parameter updating in this group. As we need to get the saliency map in time after the convolution group has been updated, we copy the parameter in convolution group to deconvolution group before we calculate the saliency map.

The saliency map is used to supervise the  AU local features, and we also take the whole image feature into consideration. The global feature is extracted from the bottleneck layer and supervised by triplet loss, and the margin is determined by the difference of pain intensity between the anchor and negative example. We combine local and global feature through a multi-task learning framework, using alternate training strategy.

Finally we finetune the network from the bottleneck layer supervised by pain intensity with regression loss, and get better result than the previous work \cite{Wang2017Regularizing}.

	In summary, the contributions of this work include:
	
1. Use triplet to build the attention map for pain regression task without sampling and datasets banlancing, which can make full use of the ill-posed datasets.

2. Supervise the saliency map with attention map so the bottleneck feature is embedded with more information of the interest area related with the pain regression task.

3. Propose a method to train the network as a multi-task problem combining local feature and global feature.

\section{Related works}
\label{sec:related_works}
\subsection{Pain Intensity Regression}

The work on pain intensity based on computer version has significant improvement due to the deep learning technology and the release of the \textit{UNBC-McMaster Shoulder-Pain}.
 There are two main streams, video based and image based methods.

Personalized method \cite{Martinez2017Personalized} uses the facial point of face image as input and uses Bi-LSTM \cite{DBLP:journals/corr/MaH16} to estimate the observed pain intensity (OPI) value. They build individual facial expressiveness score (I-FES) for each person and use Hidden Conditional Random Fields (HCRFs) to merge the sequence result to get personalized visual analog scales (VAS) estimation. Recurrent Convolutional Neural Network Regression (RCR) \cite{Zhou2016Recurrent} uses recurrent convolution network leveraging sequence information, and is trained end-to-end yet achieving sub-optimal performance. The other direction is image based method \cite{Wang2017Regularizing} that changes the face verification task to be a regression task, using smooth $L1$ \cite{Girshick2015Fast} loss and adding center loss \cite{Wen2016A} to make the result more discrete. FATAUVA-Net  presents a deep learning framework for Valence-Arousal  (V-A) estimation that employs AUs as mid-level representation where they map the AU labels to the feature maps as a surprised signal, training different branch for different AUs and using the AU detection results for Valence-Arousal (V-A) estimation.

\subsection{Regulation Loss}

Regulation restricts the size of the parameter space, so the deeper neural network can also have good generalization ability learning from small datasets. $L1$ regulation makes parameters sparsely while $L2$ weight decay restricts the norm of parameters. Recently proposed methods \cite{Miyato2015Distributional} \cite{Miyato2017Virtual} focus on small datasets training or semi-supervised learning, which are similar with ours, however, we add the regulation on saliency map. To our best knowledge, no previous work was carried in such direction.

\section{Our method}
\label{sec:our_method}

\subsection{Attention Map Generation}

Since only some of the AUs are relevant with pain intensity, we organize the inputs images in the form of triplets to use the relative values of pain intensity instead of the absolute values. Building attention map in the form of triplets, we can exclude the irrelevant AUs that wave obviously among same pain intensity or barely change among different pain intensity. Supervised by such attention map, the saliency map areas corresponding to irrelevant AUs will have almost the same brightness, which means the areas corresponding to irrelevant AUs in input image are embedded as little as possible into the bottleneck features, and the areas corresponded with relevant AUs in input images are embedded as much as possible into the final bottleneck features. 

In details, we choose the triplet:  the anchor and positive have the same pain intensity, and the negative has different. We divide all  AUs into set $\set{A}$ and set $\set{B}$. AUs in set $\set{A}$ are relevant AUs whose corresponded areas in saliency map should discriminate in a triplet,  and AUs in set $\set{B}$ are irrelevant AUs whose corresponded areas in saliency map should almost be the same in a triplet. AU $\var{k}$ is selected to set $\set{A}$ if it satisfy the following equation, otherwise it was added to set $\set{B}$. Set $\set{A}$ contains relevant AUs that barely change between anchor and positive and wave obviously between anchor and negative.

\begin{equation}
\label{equ:triplet}
    \left|V{_a^k} - V{_p^k}\right|<\alpha\quad and \quad \left|V{_a^k} - V{_n^k}\right|\geq\alpha
\end{equation}
 $V^\var{k}$ is the value of facial AU $\var{k}$. In a triplet,  we denote the value of facial AU $\var{k}$ for anchor, positive, negative as $V{_\var{a}^\var{k}}$,  $V{_\var{p}^\var{k}}$,  $V{_\var{n}^\var{k}}$ respectively. $\var{\alpha}$ is the threshold depending on the facial AU value.  
 
 Since AUs are corresponding to the semantic pattern of facial attributes, we use areas around each AU to generate the attention map, which is similar to the approach in EAC (enhancing and cropping) Net \cite{Li2017EAC}. Note that EAC Net  uses the absolute value of AU to generate the attention map for AU detection task. Different from their approach, we use the relative value of AU in a triplet for pain intensity regression as for not all AUs are related with pain intensity. We use attention map as direct signals for saliency supervision while EAC Net  adds attention map to feature map.
 \subsection{Saliency Map Supervision}
\begin{figure}[htb]
  \centering
  \centerline{\includegraphics[width=14.0cm]{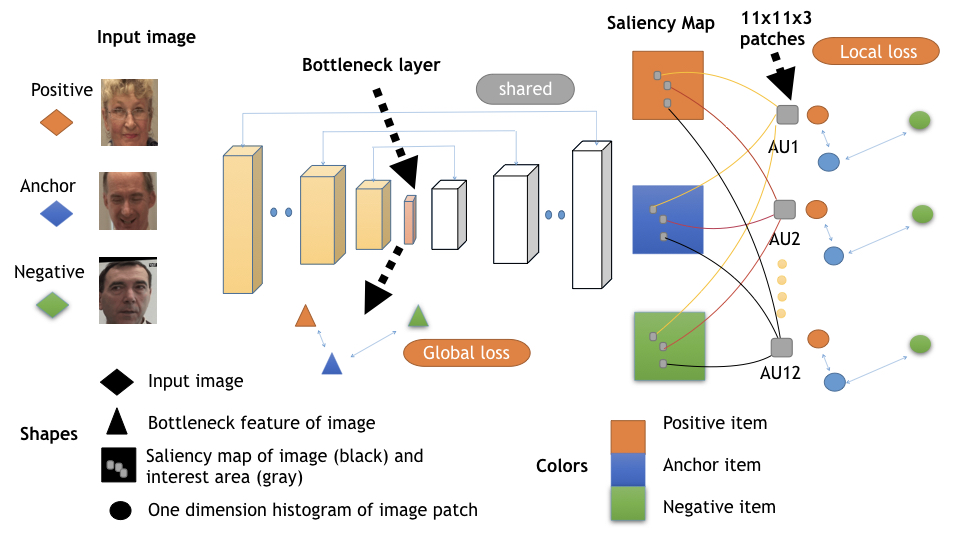}}
%
\caption{Our proposed network structure.}

\label{fig:res0}
\end{figure}


 Figure \ref{fig:res0} shows our proposed network structure. The saliency map is observed from the bottleneck feature using deconvolution network (white blocks) that has the same architecture and parameters as the convolution network (yellow blocks). For each AU, we find the corresponding patch in anchor, positive and negative item and every patch is 11 $\times$ 11 pixels around the corresponding facial landmarks. The local loss is the sum of triplet loss generated by each AU.

We divide AUs into two set $\set{A}$ and $\set{B}$ in Section 3.1. The loss function for two sets is different.  

The loss for set $\set{A}$ is:

\begin{equation}
    loss_{\set{A}} = \sum_{k\subseteq{\set{A}}}([g(P{_a^k}, P{_n^k})-g(P{_a^k}, P{_p^k})+O*W^k]_+)
\end{equation}

\begin{equation}
    W^\var{k} = \frac{\left| V{_\var{a}^\var{k}}-V{_\var{n}^\var{k}}\right|}{\sum_{i\subseteq{\set{A}}(\left|V{_\var{a}^\var{i}}-V{_\var{n}^\var{i}}\right|)}}
\end{equation}

The loss for set $\set{B}$ is:

\begin{equation}
    loss_{\set{B}} = \sum_{k\subseteq{\set{B}}}([g(P{_a^k}, P{_n^k})-g(P{_a^k}, P{_p^k})+O*\frac{1}{N}]_-)
\end{equation}

We denote saliency map patch of AU $\var{k}$ in anchor, positive, negative as $\var{P{_a^k}}$, $\var{P{_p^k}}$, $\var{P{_n^k}}$ respectively.  $\var{g(m, n)}$ is the distance metric which we will discuss in details.
$\var{O}$ is the absolute value between the pain intensity of anchor and negative. $\var{W^\var{k}}$ represents the contribution of AU $\var{k}$ to pain intensity in set $\set{A}$. $\var{N}$ is the number of AUs in set $\set{B}$, and we average the influence of each AU to pain intensity in set $\set{B}$.

We denote the distance metric by $\var{g(m, n)}$, in practical, we use normalized Earth Mover's Distance \cite{Rubner:2000:EMD:365875.365881} between the one dimensional histogram of image patch $\var{m}$ and $\var{n}$, as it is robust to face alignment error and performs better when evaluating the saliency map difference according to the image retrieval method \cite{Rubner:2000:EMD:365875.365881}.

The total local loss is:

\begin{equation}
    loss_L = loss_{\set{A}} + loss_{\set{B}}
\end{equation}

The supervision signal is saliency map, making the bottleneck feature embedded with similar information of the saliency map. To the best of our 
knowledge, we first supervise the saliency map for the origin task training.

\subsection{Multi-task Training}
The local loss we get in Section 3.2 focus on the facial AU areas which acts as the regulation loss. We find that if we only use the local loss, the network can not converge, so we combine it with the global feature together through multi-task training strategy and the local loss acts as the regulation loss. The normalized bottleneck feature is taken as the global feature presentation for the whole image, and we train it with triplets loss using Euclidean distance. The single triplet loss is as the following equation:

\begin{equation}
    loss_G = [f(P{_a}, P{_n})-f(P{_a}, P{_p})+O*\beta]_+
\end{equation}

We denote the bottleneck feature of anchor, positive, negative as $\var{P{_a}}$, $\var{P{_p}}$, $\var{P{_n}}$ respectively.  $\var{f(m, n)}$ is the Euclidean distance. $\var{O}$ is the absolute value between the pain intensity of anchor and negative and we multiply $\var{O}$ by factor $\beta$ as the triplet loss margin. During training, we carry hard negative sampling \cite{DBLP:journals/corr/HermansBL17} in all triplets in a batch, and we use alternate training strategy to combine the global loss and local loss.
   
\subsection{Pain Regression}
After training the network using local feature and global feature, we take the pain regression task. We finetune the network from the bottleneck layer, and as for the pain intensity is set to [0, 5], we use the activation function $\sigma(x) = \frac{5} {1+e{^{-x}}}$, $\var{x}$ is the output of bottleneck layer. we choose the smooth $L1$ \cite{Girshick2015Fast} + $L1$ center loss \cite{Wen2016A} as our regression loss.

\section{Experiments}
\label{sec:experiments}
	In this section we will explain our implement details and experiments result.
	
	We choose the related face recognition task and use the state of art pertained model as pretrained model. We use MTCNN \cite{DBLP:journals/corr/ZhangZL016} to detect the face area, and do face alignment according to the facial landmarks provided by \textit{UNBC-McMaster Shoulder-Pain} dataset. The AU labels are available from \textit{UNBC-McMaster Shoulder-Pain} dataset.

We train the attention network alternatively using basic learning rate 0.001 for global loss, and 0.01 for local loss. When we train the local loss, the deconvolution group is frozen, and share the parameters with the convolution group. During training we first select hard negative examples \cite{DBLP:journals/corr/HermansBL17} in a batch, and train the network using $loss_G$, and then we use the same hard negative examples to get the $loss{_L}$ to train the network again.

When training  the pain intensity regression, we set pain intensity value from [0, 15] to [0, 5] as proposed by the previous work\cite{Zhao2016Facial} \cite{Zhou2016Recurrent}. We evaluate the result using 25-cross-valuation. We try to train the network only using the local loss, however the network is not converging, and when the global loss is added, the network performance well which means the local loss acts like regulation loss here. 

\begin{figure}[htbp]
	\centering
	\subfigure[]{
	\begin{minipage}{2.5cm}
	\centering
		\includegraphics[width=2.5cm]{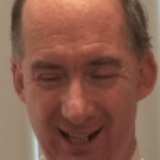}
	\end{minipage}
	}	
	\subfigure[]{
	\begin{minipage}{2.5cm}
	\centering
		\includegraphics[width=2.5cm]{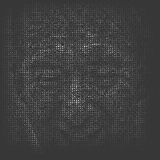}
	\end{minipage}
	}
	\subfigure[]{
	\begin{minipage}{2.5cm}
	\centering
		\includegraphics[width=2.5cm]{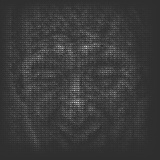}
	\end{minipage}
	}	
	\subfigure[]{
	\begin{minipage}{2.5cm}
	\centering
		\includegraphics[width=2.5cm]{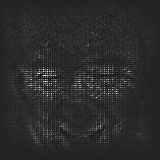}
	\end{minipage}
	}	
\caption {Saliency maps of different network training approaches. (a) is the input picture, (b) is the saliency map from bottleneck embeddings trained directly by the pain intensity, (c) is the saliency map from bottleneck embeddings trained by the global feature and finetuned by the pain intensity, (d) is the saliency map from bottleneck embedding trained by the combination of global feature and local feature and finetuned by pain intensity. }
\label{fig:res1}
\end{figure}

	We compare the difference of directly taking regression for the task, using global feature then finetune the network for regression, combining global feature and local feature then finetune the network for regression. The comparsion of saliency map can be found in Figure \ref{fig:res1}. The experiment results can be found in Table 1 and Table 2.

Brighter areas in the saliency map means that the corresponded areas in input image are more embedded into the bottleneck feature. As we can see from Figure \ref{fig:res1}, using global feature and then finetuned by the pain intensity, the model focuses some area of face while not accurate enough, when using local feature and global feature together, the model focuses on the eye, mouth areas that related to the pain expression.

\begin{table}[htb]
\label{tab:res1}
\begin{center}
    \begin{tabular}{ | l | l | l | l|}
    \hline
    Methods & MAE$\downarrow$ & MSE$\downarrow$ & PCC$\uparrow$ \\ \hline
    our proposed method 1 & 0.401 & 0.742 & 0.643 \\ \hline
    our proposed method 2 & {\bfseries0.334} & {\bfseries0.626} & {\bfseries0.804}   \\ \hline
    smooth $L1$ + $L1$ center loss\cite{Wang2017Regularizing}   & 0.456 & 0.804 & 0.651 \\    \hline
    OSVR-$L1$\cite{Zhao2016Facial}               & 1.025 & N/A & 0.600\\    \hline
    OSVR-$L2$\cite{Zhao2016Facial}               & 0.810 & N/A & 0.601\\    \hline
    RCR\cite{Zhou2016Recurrent}                   & N/A & 1.54 & 0.65\\    \hline
    \end{tabular}
\end{center}

\caption{Performance of our  proposed methods and related works on the \textit{UNBC-McMaster Shoulder-Pain}  dataset for the estimation of pain intensity. Our proposed method 1: Use global feature to train the network and smooth $L1$ + $L1$ center loss for regression.  Our proposed method 2: Use the combination of global feature and local feature to train the network and  smooth $L1$ + $L1$ center loss for regression. MAE is short for mean absolute error deviated from the ground-truth labels over all frames per video. MSE is mean squared error which measures the curve fitting degree. PCC is Pearson correlation coefficient which measures the curve trend similarity ( $\uparrow{}$ indicates the larger, the better). The best is highlighted in bold.}

\end{table}

\begin{table}[htb]
\label{tab:res2}
\begin{center}
    \begin{tabular}{ | l | l | l |}
    \hline
    Methods & wMAE$\downarrow$ & wMSE$\downarrow$\\ \hline
    our proposed method 1 & 0.883 & 1.697\\ \hline
    our proposed method 2 & {\bfseries0.727} & {\bfseries1.566} \\ \hline
    smooth $L1$ + $L1$ center loss + sampling\cite{Wang2017Regularizing}   & 0.991 & 1.720 \\    \hline
    OSVR-$L1$\cite{Zhao2016Facial}               & 1.309 & 2.758\\    \hline
    OSVR-$L2$\cite{Zhao2016Facial}               & 1.299 & 2.719 \\    \hline
    \end{tabular}
\end{center}

\caption{Performance of our network when evaluated using the weighted MAE and weighted MSE proposed by \cite{Wang2017Regularizing}. Methods proposed by us are applied with uniform class sampling technique.}
\end{table}

\section{Summary}
\label{sec:summary}

In this paper, we propose a novel method for attention based network. We use deconvolution network to get the saliency map of the bottleneck layer, and design the attention map for pain intensity regression task. Through direct regulation on saliency map and multi-task training with global loss, we successfully push forward single image based method by a considerable margin without bells and whistles.
We believe our job is great for single image based prediction, and the performance of joint prediction over image sequences should benefit from features extracted by our method, which we leave as a future direction.

\bibliographystyle{IEEEbib}
\bibliography{strings,refs}
\end{document}